\documentclass[12pt]{article}
\usepackage{arxiv}
\usepackage{graphicx,epstopdf}
\usepackage{amsmath}
\usepackage{times}
\usepackage{graphicx}
\usepackage{color}
\usepackage{multirow}
\usepackage{mathtools}
\usepackage{amssymb}
\usepackage{booktabs}       
\usepackage{amsfonts}       
\usepackage{nicefrac}       
\usepackage[authoryear]{natbib}	

\title{A sparse code increases the speed and efficiency of neuro-dynamic programming for optimal control tasks with correlated inputs}

\date{} 					

\author{ Peter N.~Loxley\\
	School of Science and Technology,\\
	University of New England,\\ 
	Armidale 2351, NSW, Australia.
}

\renewcommand{\headeright}{A Postprint}
\renewcommand{\undertitle}{A Postprint}

\begin{document}
\maketitle

\begin{abstract}
Sparse codes in neuroscience have been suggested to offer certain computational advantages over other neural representations of sensory data. To explore this viewpoint, a sparse code is used to represent natural images in an optimal control task solved with neuro-dynamic programming, and its computational properties are investigated. The central finding is that when feature inputs to a linear network are correlated, an over-complete sparse code increases the memory capacity of the network in an efficient manner beyond that possible for any complete code with the same-sized input, and also increases the speed of learning the network weights. A complete sparse code is found to maximise the memory capacity of a linear network by decorrelating its feature inputs to transform the design matrix of the least-squares problem to one of full rank. It also conditions the Hessian matrix of the least-squares problem, thereby increasing the rate of convergence to the optimal network weights. Other types of decorrelating codes would also achieve this. However, an over-complete sparse code is found to be approximately decorrelated, extracting a larger number of approximately decorrelated features from the same-sized input, allowing it to efficiently increase memory capacity beyond that possible for any complete code: a 2.25 times over-complete sparse code is shown to at least double memory capacity compared with a complete sparse code using the same input. This is used in sequential learning to store a potentially large number of optimal control tasks in the network, while catastrophic forgetting is avoided using a partitioned representation, yielding a cost-to-go function approximator that generalizes over the states in each partition. Sparse code advantages over dense codes and local codes are also discussed.
\end{abstract} 
 
\section{Introduction}\label{introduction}
Sparse codes have traditionally been viewed as efficient, low bit-rate representations of sensory data such as natural images \citep{barlow, daugman89, field, hyvarinenbook}. Evidence also exists that the mammalian visual cortex makes use of a sparse code for visual stimuli such as natural images \citep{gallant}. It has long been known that pixel representations of images are highly redundant (see historical references in \cite{petrov}, and \cite{eichhorn}, for example). Spatial correlations mean the value of a pixel at one point often leads to a reasonable prediction for the value of a pixel at a nearby point, and higher-order statistical dependencies are also present. Daugman showed it is possible to transform an image to a new representation with fewer statistical dependencies and a lower entropy which he termed a \emph{sparse code} \citep{daugman89}. This was originally done using a complete, discrete, two-dimensional Gabor transform \citep{daugman88}. A well-known property of the two-dimensional (2D) Gabor function is that its parametric form also describes neural receptive field profiles \citep{daugman85,jones}. Olshausen and Field later demonstrated the converse of Daugmans' work, namely that learning a sparse code for natural images leads to profiles similar to those of neural receptive fields \citep{olshausen96,olshausen97}. More recent work relates to finding highly over-complete sparse codes \citep{sommer,olshausen13b}, and the role of homeostasis during the process of learning a sparse code \citep{perrinet1,perrinet2}.

If sparse coding is a neural coding principle  then it should have computational advantages over other coding approaches. F{\"o}ldi{\'a}k originally compared sparse codes with local codes and dense codes (natural images correspond to dense codes in this work). Suggested advantages included a high memory capacity, a fast speed of learning, controlled interference between different stored patterns, high fault tolerance (relative to local codes), and a high representational capacity \citep{fold}. These suggestions drew from work on associative memories and neural networks \citep{hertz}, rather than from information theory. In addition to the possibility of sparse coding as a neural coding principle , recent successes in the field of deep reinforcement learning \citep{mnih} make these suggestions worth re-visiting. The reason is the success of deep learning could very well be due to sparse codes, as suggested recently by \cite{papyan}. It is therefore worthwhile establishing the fundamental computational properties of a sparse code.

The aim of this work is to investigate the computational properties of a sparse code starting with F{\"o}ldi{\'a}k's suggestions, and to determine what advantages a sparse code may have for solving optimal control tasks using neuro-dynamic programming. The central finding is that the combination of decorrelation and over-completeness gives a sparse code a computational advantage over other codes in optimal control tasks solved with neuro-dynamic programming. In this respect, the present work both clarifies and extends F{\"o}ldi{\'a}k's suggestions for the computational advantages of a sparse code. The optimal control task considered here involves tracking a target object, given by a dragonfly, over a sequence of temporally-correlated natural images. Each image in the sequence can be represented as a sparse code and used in a linear network as a function approximator. Neuro-dynamic programming can then be applied to solve the optimal control task. The sparse code is generated using a recently developed method for constructing a suitable basis of 2D Gabor functions that is scale-invariant and adapted to natural image statistics \citep{loxley}. The advantage of this approach is that all computational problems addressed in this paper can then be solved using either dynamic programming \citep{bert}, or convex optimization \citep{boyd}. The primary aim of this work is to investigate sparse code properties that generalize to any optimal control task, not just specific tracking tasks. However, a proposed non-greedy online tracking algorithm is briefly discussed in Section \ref{conclusion}.

The structure of the paper is as follows. Section \ref{model} introduces the dynamic programming model and describes how the sparse codes are generated. Details of the optimal control task are given in Section \ref{results}, and results are presented for memory capacity, speed of learning, sequential learning of multiple tasks, representational capacity, and fault tolerance. A summary of the main findings appears in Section \ref{conclusion}, and possible extensions are suggested.

\section{Model}\label{model}
Consider the optimal control problem of tracking a target over a temporal sequence of $N$ images. Rather than detecting the location of the target in each image, the task is to follow the known location of the target as closely as possible by applying a restricted set of discrete controls. The tracking task is therefore a combinatorial optimization problem, and suboptimal solutions will generally be present. 

Let the state $x_k$ be a pair of coordinates giving the location of a small region $R_k(x_k)$ within the $k$th image of the image sequence, and let $w_k$ be the known location of the target region $R_k(w_k)$. To be precise; $x_k, w_k\in\mathbb{Z}^2$, and let $R_k$ be an $a\times a$ pixel region represented by the vector $R_k\in\mathbb{R}^q$ (where $q=a^2$). The parameter $a$ is chosen according to some relevant lengthscale of the target object, such as its maximum length in pixels in the sequence of images. The tracking dynamics is assumed to be deterministic, and is given by the discrete-time dynamical equation:
\begin{equation}
x_{k+1}=x_{k} + u_{k},\label{td}
\end{equation}
where $u_k$ is the control applied at each stage to update the state in the next image in the sequence. The key feature of this model is that the set of controls $\{u_k\}_{k=1}^N$ is restricted: each $u_k\in U$ is taken from the set 
\begin{equation}
U=\{(-a,0), (a,0), (0,a), (0,-a), (0,0) \},
\end{equation}
corresponding to a shift of $a$ pixels either left, right, up, down, or no shift if $u_k=(0,0)$. Restricting controls makes the tracking problem more interesting, but also more difficult to solve, than for the case of continuous controls. In some tracking situations it might be practically relevant to consider controls that are restricted in certain ways. The dynamical equation (\ref{td}) implies the maximum distance travelled by the target from one image to the next must be $a$ pixels. Targets moving faster than this will outrun the tracker unless image sequences are resampled at a higher frequency. The total cost for the tracking problem is assumed to take the simple form:
\begin{equation}
\sum_{k=1}^{N}|x_k-w_k|^{2},\label{tc}
\end{equation}
penalizing all deviations from the target location accumulated over the $N$-image sequence. 

Minimizing the total cost given by (\ref{tc}) will result in an optimal tracking solution. This can be done exactly using the \emph{Dynamic Programming (DP) algorithm} \citep{bert}. Assuming a terminal cost of $J_{N+1}(x_{N+1})=0$, the DP algorithm involves iterating backwards from $J_{N}(x_{N})$ to $J_{1}(x_1)$ using the following equation:
\\
\begin{equation}
J_{k}(x_{k})=\underset{u_k\in U}{\operatorname{min}}\left[|x_k-w_k|^{2}+J_{k+1}(x_{k}+u_k)\right],\label{dp}
\end{equation}
where $J_{k}(x_{k})$ is the \emph{cost-to-go}: giving the tail portion of the cost remaining when going from state $x_{k}$ and image $k$, to state $x_N$ and image $N$, using the optimal control sequence. The minimum total cost is then given by the value of $J_{1}(x_1)$, and the corresponding choice for each $u_k$ gives the set of optimal controls $\{u_k^*\}_{k=1}^N$ \citep{bert}. Using these optimal controls in equation (\ref{td}) returns the optimal tracking solution for a given temporal sequence of images. A simple one-dimensional tracking problem is solved in the appendix following this approach, and solutions are shown in Figures 1 and 3 for the more complicated tracking task described in Section 3.

\subsection{Neuro-dynamic programming}
There are two immediate problems with the exact DP approach outlined above. The first is that when dealing with images it is often the case that choosing a suitable representation can make a task easier to solve. The second is that the table of cost-to-gos $J_{k}(x_{k})$ may become prohibitively large, especially if the number of states increases exponentially with problem size. Both issues can be addressed by replacing $J_{k}(x_{k})$ with a parametric function approximator $\tilde{J}_{k}(x_{k},r_k)$. The function approximator could be given by a neural network, for example. The parameters $r_k$  are usually called weights and can be found using \emph{fitted value iteration} \citep{bert,bert2} as follows. Given some sample states $x_{k}^0,x_{k}^1,...$ for each $k$,  this algorithm involves iterating backwards from $\tilde{J}_{N}(x_{N}^s,r_{N})$ to $\tilde{J}_{1}(x_{1}^s,r_1)$ using the following pair of expressions:
\begin{align}
&\beta_k^s=\underset{u_k\in U}{\operatorname{min}}\left[|x_k^s-w_k|^{2}+\tilde{J}_{k+1}(x_{k}^s+u_k,r_{k+1})\right],\label{dpiter}\\
&\underset{r_k}{\operatorname{min}} \sum_s |\tilde{J}_{k}(x_{k}^s,r_k)-\beta_k^s |^{2}.\label{lsfit}
\end{align}
The equation (\ref{dpiter}) performs a DP iteration similar to equation (\ref{dp}) to find the cost $\beta_k^s$ corresponding to sample $x_{k}^s$. This state-cost sample pair $(x_k^s,\beta_k^s)$ is then used in (\ref{lsfit}) as a training sample to fit $\tilde{J}_{k}(x_{k}^s,r_k)$ to its target value $\beta_k^s$ by adjusting the weights $r_k$. A simple form for $\tilde{J}_{k}(x_{k},r_k)$ is given by a linear network:
\begin{equation}
\tilde{J}_{k}(x_{k},r_k)=r_k^Tv_k(x_k),\label{nn}
\end{equation}
where $r_k\in\mathbb{R}^p$ is a vector of weights, and $v_k(x_k)\in\mathbb{R}^p$ is a vector of network inputs given by features extracted from image region $R_k(x_k)\in\mathbb{R}^q$. In the case of natural images, the raw image pixels are features that are strongly correlated. Alternative feature vectors (image representations) may make an optimal control task easier to solve. Image representations may either be \emph{complete:} $p=q$, \emph{over-complete:} $p>q$, or \emph{under-complete:} $p<q$. With this choice of function approximator, equation (\ref{lsfit}) reduces to linear regression and the corresponding minimization of a convex quadratic function \citep{bert2, boyd}. One approach to solving equation (\ref{lsfit}) that is easy to implement is the \emph{incremental gradient method} \citep{bert2}. For the $k$th image in the sequence, solving equations (\ref{lsfit}) and (\ref{nn}) with the incremental gradient method requires updating the weights $r_k$ according to
\begin{equation}
r^{(t+1)}_k = r_k^{(t)} - \eta \left(v_k(x^s_k)^Tr_k^{(t)}-\beta^s_k\right)v_k(x^s_k).\label{grad}
\end{equation}
This update is performed for each pair of state-cost samples $(x_k^s,\beta_k^s)$. By cycling through all sample pairs for the $k$th image, and then repeating a number of times, convergence is guaranteed provided the learning rate $\eta$ is chosen carefully. The algorithm for \emph{incremental value iteration} \citep{bert2} combines equations (\ref{dpiter}) and (\ref{nn}) with (\ref{grad}), and is the algorithm of choice for neuro-dynamic programming (neuro-DP) in this investigation. When the number of states of a problem is small enough to compute the exact cost-to-go $J_{k}(x_{k})$, this can be compared with $\tilde{J}_{k}(x_{k},r_k)$ from neuro-DP to determine the quality of the approximation. 

\subsection{A sparse code for natural images}
It now remains to choose a suitable image representation to approximate the cost-to-go in equation (\ref{nn}). This representation should help make the tracking task easier to solve in some way, and it is the purpose of the present investigation to determine how this might be done using sparse codes.

A simple generative model described in \cite{loxley} will now be used to find a sparse code for natural images. A sampling scheme approximating the joint probability density of parameter values of the 2D Gabor function adapted to natural image statistics is reproduced in Table 1.
\begin{table}
\centering
\begin{tabular}{ll}
\hline
Gabor Parameter(s)& Sample Transformation\\
\hline
$\sigma_x^\prime,\sigma_y^\prime,\lambda^\prime$&$(\sigma_x^\prime,\sigma_y^\prime,\lambda^\prime)=(1,1,\rho)z$\\
$\sigma_x$& $\sigma_x={\cal{PCDF}}^{-1}({\cal{NCDF}}(\sigma_x^\prime|0,1)|\alpha_1,\beta_1)$\\
$\sigma_y$& $\sigma_y={\cal{PCDF}}^{-1}({\cal{NCDF}}(\sigma_y^\prime|0,1)|\alpha_2,\beta_2)$\\
$\lambda$& $\lambda={\cal{PCDF}}^{-1}({\cal{NCDF}}(\lambda^\prime|0,1)|\alpha_3,\beta_3)$\\
\hline
\end{tabular}
\vspace{10pt}
\caption{Sampling scheme for the three spatial Gabor function parameters: sample $z\sim {\cal{N}}(0,1)$ from the standard normal distribution, then apply the parameter transformations listed in the table. Here, $\rho, \alpha_i$, and $\beta_i$ are the model parameters, ${\cal{PCDF}}^{-1}(x|\alpha,\beta)=\frac{\beta}{(1-x)^{1/\alpha}}$ is the inverse CDF for the Pareto distribution, and ${\cal{NCDF}}(x|0,1)$ denotes the CDF for the standard normal distribution \citep{loxley}.}
\end{table} 
The three spatial Gabor parameters $\sigma_x,\sigma_y,$ and $\lambda$ are strongly correlated and have heavy-tailed distributions which can be modelled using a Gaussian copula with Pareto marginal distributions. Other Gabor parameters are sampled uniformly over their respective ranges and are not shown in Table 1.In the first step, a sample is collected for each of the seven Gabor parameters: $(\phi,\varphi,\sigma_{x},\sigma_{y},\lambda,x_0,y_0)$, and a real-valued 2D Gabor function is constructed from these samples using the following equations: 
\begin{equation}
G({r},{r}^{\prime})=A\exp{\left[-\frac{1}{2}\left(\frac{\tilde{i}^2}{\sigma_{x}({r}^{\prime})^2}+\frac{\tilde{j}^2}{\sigma_{y}({r}^{\prime})^2}\right)\right]}\cos{\left[k({r}^{\prime})\tilde{j}+\varphi({r}^{\prime})\right]},\label{dic1}
\end{equation}
and
\begin{equation}
(\tilde{i},\tilde{j})=\left(\begin{array}{cc}
\cos{\phi({r}^{\prime})}&-\sin{\phi({r}^{\prime})}\\
\sin{\phi({r}^{\prime})}&\cos{\phi({r}^{\prime})}\\
\end{array}
\right)
\left(\begin{array}{c}
i-x_0({r}^{\prime})\\
j-y_0({r}^{\prime})\\
\end{array}
\right),\label{dic2}
\end{equation}\\
with $k({r}^{\prime}) = 2\pi/\lambda({r}^{\prime})$. One set of seven parameter samples corresponds to a single 2D Gabor function, which is indexed by a single value of ${r}^{\prime}$ in equations (\ref{dic1}) and (\ref{dic2}). Repeating this step $m$ times leads to $m$ Gabor functions (and $m$ values of ${r}^{\prime}$), which are now summed together to give the image model: $\hat{I}({r})=\sum_{r^{\prime}=1}^{m}G({r},{r}^{\prime})a({r}^{\prime})$. Here, ${r}=(i,j)$ are pixel coordinates of the generated image, and ${r}^{\prime}=(i',j')$ are discrete coordinates of the sparse code formed by the coefficients $a({r}^{\prime})$, each one corresponding to a particular 2D Gabor function. The sampling scheme in Table 1 is length-scale invariant due to the Pareto marginal distributions. Scale invariance is a key property of natural images \citep{ruderman2,ruderman,mumford}, and also of the underlying joint probability density approximated using Table 1 \citep{loxley}. The set of randomly generated Gabor functions are therefore self-similar and multiscale in a manner comparable to a self-similar multiresolution wavelet scheme.

Two-dimensional Gabor functions are not orthogonal. However, given an image $I({r})$, it is possible to find its sparse code $a({r}^{\prime})$ using a least-squares approximation \citep{daugman88}. Due to algorithm efficiency issues it is easiest to divide each image region $R_k(x_k)$ into a number of smaller regions $I\in\mathbb{R}^d$ with $d\ll q$. Letting $G\in\mathbb{R}^{d\times m}$ be a matrix with elements $G({r},{r}^{\prime})$, and $a\in\mathbb{R}^m$ be a vector of the sparse code coefficients $a(r^{\prime})$, the sparse code is found by solving the least-squares problem,
\begin{equation}
\begin{aligned}
&\underset{a}{\operatorname{min}} & & |Ga - I |^{2}.\label{gabor}
\end{aligned}
\end{equation}
This problem is convex and can be efficiently solved with standard solvers. An $l_1$ regularization term can also be added to Equation (\ref{gabor}) as shown in the Appendix. However, for the 2D Gabor function basis used here this did not lead to any increase in sparse code performance.

\section{Results}\label{results}
The model developed in the previous section is now applied to the optimal control task of tracking a dragonfly over a temporal sequence of natural images taken from video. The problem formulation allows for easy detection of failure in either the DP method or the approximation method. Failure in the DP method occurs when the optimal solution cannot be found. As previously discussed, applying a restricted set of discrete controls during tracking leads to a combinatorial optimization problem with suboptimal solutions. The existence of suboptimal solutions is demonstrated in the appendix for a simplified problem, and similar ideas generalize to the present case. These suboptimal solutions can usually be found with a greedy algorithm. Failure of the DP method can therefore be detected when the DP solution matches the greedy solution (unless the greedy solution also happens to be optimal). Failure in the approximation method can be detected by comparing neuro-DP with exact DP though either the costs, or the cost-to-gos. The problem formulation also renders the use of state-space models such as the Kalman filter (often used in tracking) unnecessary. The reason is the position of the target is already known precisely, rather than having to be inferred from a noisy observation. 

An image sequence and a solved tracking task is shown in Figure 1. 
\begin{figure}
\centering
\includegraphics[scale=0.9, bb = 0 34 419 311, clip = true]{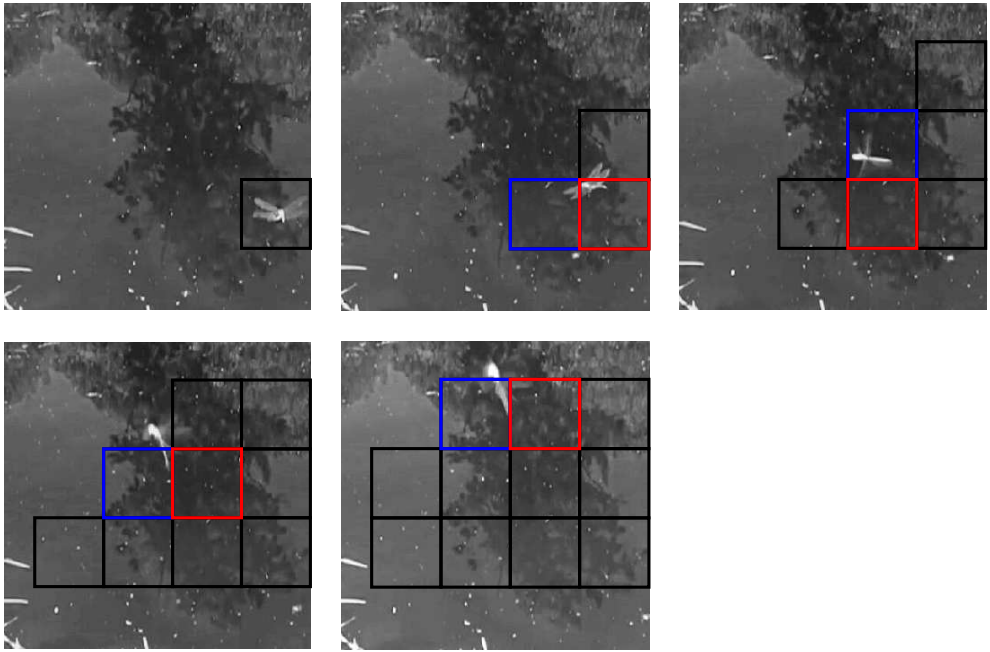}
\caption{A solved dragonfly tracking task. The image sequence starts with the top-left image, and ends with the bottom-right image. The DP tracker (blue squares) uses the restricted controls available to follow the dragonfly as closely as possible over the whole image sequence. The greedy tracker (red squares) tries to get as close as possible to the dragonfly in each image.}
\end{figure} 
Image sequences were sampled from video taken at Karrawirra Parri (the Torrens River) in South Australia. In this case a dragonfly moves from the right edge of the first image, to the top edge of the last image (going from top-left to bottom-right). For each image, the target location $w_k$ was found, and the image converted to double-precision grayscale to allow for numerical calculations: no further pre-processing was done. In Figure 1, states corresponding to small image regions (shown as black squares) were generated as follows. The initial state $x_1$ in Image 1 (top-left image) is the target location $w_1$. Applying equation (\ref{td}) for each of the five controls generates the states $x_2$ shown in Image 2 (top-second-from-left). Instead of five additional states, there are only three states that fit within the image: the tracker can either move up, move left, or stay where it is. This process is repeated for each subsequent image and state; until by Image 5, eleven unique states have been generated. To solve the tracking problem with exact DP requires starting at the last image (Image 5) and working backwards. First, $J_{5}(x_{5})$ is evaluated for each of the eleven states in Image 5 using equation (\ref{dp}) with $J_{6}(x_{6})=0$. Then, $J_{4}(x_{4})$ is evaluated for each of the nine states in Image 4 using equation (\ref{dp}) with $J_{5}(x_{5})$. Continuing in this manner, and iterating equation (\ref{dp}) backwards in time, eventually leads to the initial state in Image 1 and returns the set of optimal controls. It is now possible to proceed forwards in time from Image 1 to Image 5 and solve the tracking task using the optimal controls in equation (\ref{td}). This (DP) solution is given by the sequence of blue squares in Figure 1 and is guaranteed to minimize the total cost given by (\ref{tc}). The greedy solution is given by the sequence of red squares and minimizes the cost of the current stage only. Following the greedy policy allows the target to get ahead of the tracker, which is then never able to catch up using the restricted set of controls. By contrast, a tracker following the DP policy pays a small initial cost to get ahead of the target, and is then able to keep up at later stages.

In Figure 2, the cost-to-go of the states in Image 2 (Figure 1) are shown. 
\begin{figure}
\centering
\includegraphics[scale=0.9, bb = 49 66 382 238, clip = true]{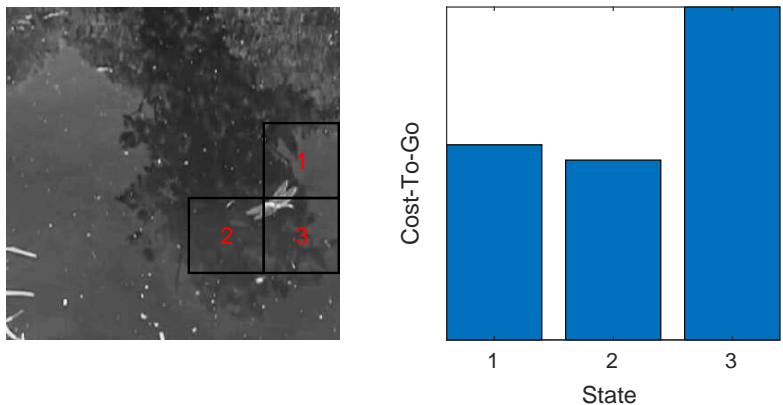}
\caption{The cost-to-go of states in Image 2 of Figure 1. State 3 is closest to the target dragonfly but has the largest cost-to-go.}
\label{}
\end{figure} 
The decision made at this stage completely distinguishes the DP policy from the greedy policy. From Figure 2, it is clear that State 3 has the largest cost-to-go. On the other hand, State 3 is closest to the target and minimizes the single-stage cost $|x_2-w_2|^{2}$ for Image 2. Therefore, the greedy tracker selects State 3, while the DP tracker chooses either States 1 or 2: either state can achieve the subsequent DP state shown in Image 3, and therefore minimizes the total cost given by equation (\ref{tc}).

Another image sequence and solved tracking task is shown in Figure 3.
\begin{figure}
\centering
\includegraphics[scale=1, bb = 0 34 419 311, clip = true]{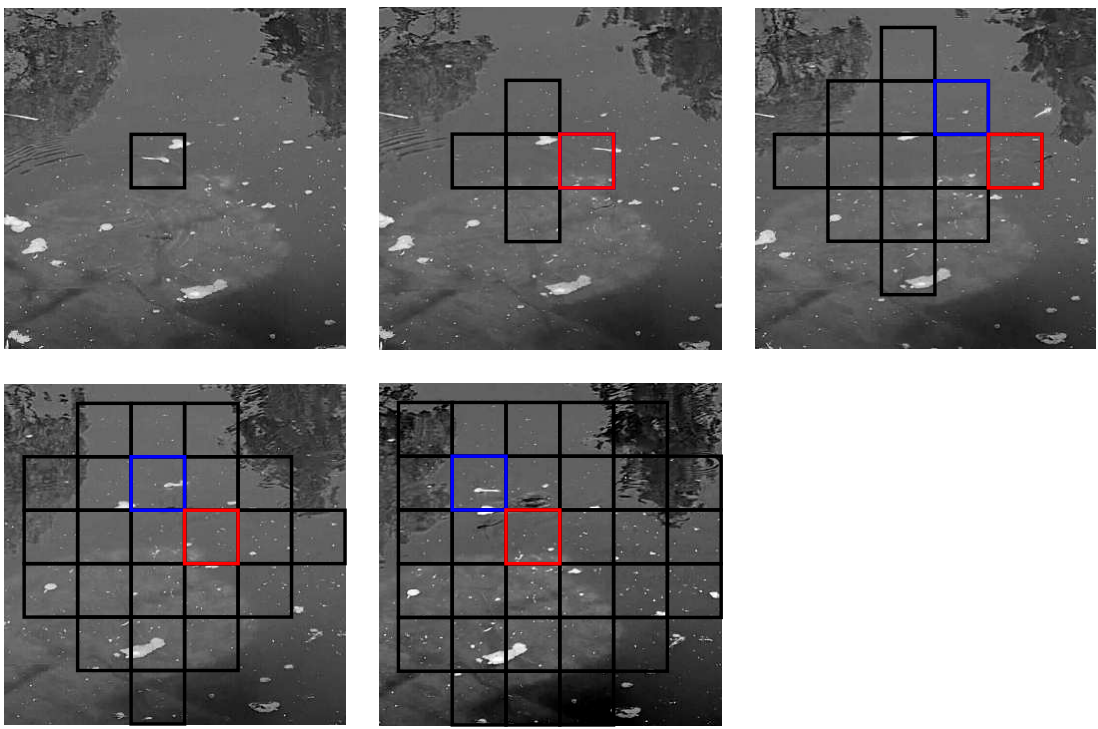}
\caption{Another solved dragonfly tracking task. The DP tracker (blue squares) and greedy tracker (red squares) both choose the same first control (second image), but different second controls (third image) as the greedy tracker stays closest to the dragonfly in the third image.}
\end{figure} 
In this example, the DP and greedy trackers both choose the same first control in Image 2 (shown by the red square). In Image 3, the DP and greedy trackers diverge: the greedy tracker moves to the state closest to the target (red square) to minimize the single-stage cost for Image 3, while the DP tracker moves to a different state to minimize the total cost given by (\ref{tc}). The cost-to-go for Image 3 is shown in Figure 4. From the thirteen possible states, State 4 has the smallest cost-to-go and is chosen by DP. However, State 9 minimizes the single-stage cost $|x_3-w_3|^{2}$ for Image 3 and is therefore the greedy choice. 
\begin{figure}
\centering
\includegraphics[scale=0.9, bb = 60 76 448 274, clip = true]{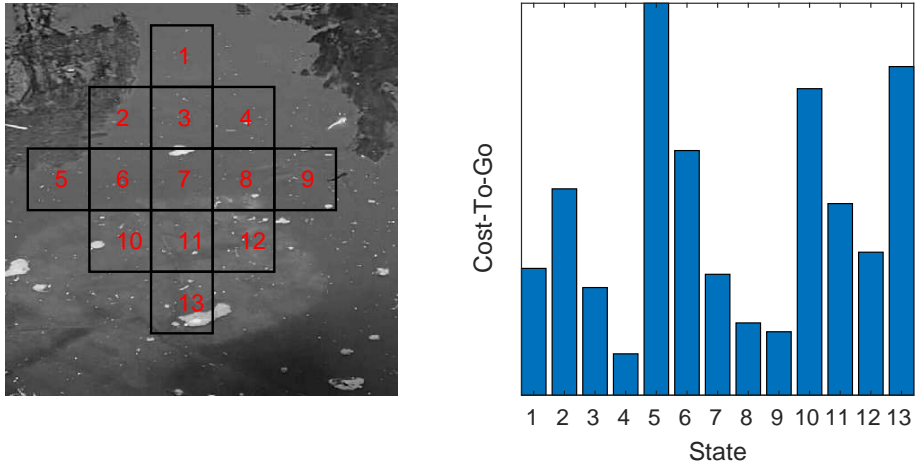}
\caption{The cost-to-go of states in Image 3 of Figure 3. State 9 is closest to the target dragonfly, while State 4 has the lowest cost-to-go.}
\label{}
\end{figure} 
From Figure 3 it is clear the DP solution, rather than the greedy solution, tracks the dragonfly most closely over the whole image sequence. 

Image representations are included using the neuro-DP framework with incremental value iteration. Choosing $v_k(x_k^s)$ in equation (\ref{nn}) to be a sparse code representation of image region $R_k(x_k^s)$, and applying equations (\ref{dpiter}), (\ref{nn}), and (\ref{grad}), leads directly to the cost-to-gos and optimal controls shown in Figures 1--4. The function used for cost-to-go approximation is a linear network, so performance on optimal control tasks like tracking will strongly depend on network properties such as memory capacity, speed of learning, the ability of the network to store multiple tasks with minimal interference between different tasks, the degree of fault tolerance within the network, as well as the capacity of the network to represent different states. These network properties are now investigated. 

\subsection{Memory capacity}
A memory can be considered as the learned association between each pair of state-cost samples $(x_k^s,\beta_k^s)$ in the least-squares regression in equation (\ref{lsfit}). Given a state sample, memory retrieval then requires accurate recall of the best approximation of the corresponding cost sample. This does not include associations due to generalization (interpolating between learned state-cost samples). Memory capacity is defined here to be the maximum number of state-cost sample associations that can be directly stored in a linear network. Once memory capacity is reached no new tasks can be learned without losing previously learned tasks. It has been suggested that memory capacity increases going from a less-sparse (dense) code to a sparser code \citep{fold}. I will now demonstrate this for the case of a linear network storing state-cost sample associations. It might be expected that a linear network with $p$ weights could reliably store $p$ cost samples, giving memory capacity $n_{\text{max}}=p$. However, this is generally not the case, and therefore a more careful analysis is required.

Making use of equation (\ref{nn}) and dropping the $k$-dependence, the objective in equation (\ref{lsfit}) can be re-written as
\begin{equation*}
\underset{r_k}{\operatorname{min}}\sum_{s=1}^n |\tilde{J}_{k}(x_{k}^s,r_k)-\beta_k^s |^{2}=\underset{r}{\operatorname{min}}|Vr-\beta|^{2},
\end{equation*} 
where $V$ is the \emph{design matrix}:
\begin{align*}
V=
\begin{bmatrix}
v_1(x^1) &\dots &v_p(x^1)\\ \vdots & &\vdots\\
v_1(x^n) &\dots &v_p(x^n)
\end{bmatrix},
\end{align*}
$r=(r_1,...,r_p)$ is a vector of network weights, and $\beta=(\beta^1,...,\beta^n)$ is a vector of cost samples. Each row of $V$ corresponds to one of $n$ samples, and each column of $V$ corresponds to one of $p$ measurements performed on each sample. So row ``$s$" gives the vector $v(x^s)$ corresponding to the representation of image region $R(x^s)$ for state sample $x^s$. In the present case, each column corresponds to a different pixel value, or a different Gabor coefficient value (there are $p$ columns for an image region of $p$ pixels or a sparse code of $p$ Gabor coefficients). The columns of $V$ span a vector subspace (the column space of $V$), and a vector in this subspace can be represented as 
\begin{align*}
Vr&=
r_1\begin{bmatrix}
v_1(x^1) \\ \vdots \\
v_1(x^n)
\end{bmatrix}+ \dots +r_p\begin{bmatrix}
v_p(x^1) \\ \vdots \\
v_p(x^n)
\end{bmatrix}.
\end{align*}
Minimizing the objective in equation (\ref{lsfit}) with respect to $r$ corresponds to projecting the vector $\beta$ onto the closest vector in the column space of $V$: $\hat{\beta}=V\hat{r}$, called the \emph{best approximation} to $\beta$. According to the definition of memory capacity, $n_{max}$ is equal to the dimension of the column space (the \emph{rank}) of $V$. If the columns of $V$ are orthogonal, then $V$ has full rank, and $n_{max}=p$. However, if any of these columns are linearly dependent it must be the case that $n_{max}<p$. To investigate further, consider the expression for the normalized inner product of any two column vectors of $V$,
\begin{equation*}
\frac{\langle v_i, v_j\rangle}{|v_i||v_j|}=\frac{\sum_s v_i(x^s)v_j(x^s)}{\sqrt{\sum_s v_i(x^s)^2}\sqrt{\sum_s v_j(x^s)^2}},\label{orth}
\end{equation*}
\vspace{5pt}\\
and the expression for the sample correlation between two random variables $v_i$ and $v_j$,
\begin{equation*}
\text{corr}(v_i,v_j)=\frac{\sum_s (v_i(x^s)-\bar{v}_i)(v_j(x^s)-\bar{v}_j)}{\sqrt{\sum_s(v_i(x^s)-\bar{v}_i)^2}\sqrt{\sum_s(v_j(x^s)-\bar{v}_j)^2}},\label{corr}
\end{equation*}
\vspace{2pt}\\
where $\bar{v}_i$ is the mean of $v_i$. In the limit of vanishing mean and constant $|v_i|$, it is clear that $\langle v_i, v_j\rangle\propto\text{corr}(v_i,v_j)$. This correspondence between $\langle v_i, v_j\rangle$ and $\text{corr}(v_i,v_j)$ allows the orthogonality expression $\sum_s v_i(x^s)v_j(x^s)$ to be qualitatively investigated using sample scatter plots, and shows that the memory capacity of a linear network only depends on the second-order statistics of a representation.

Memory capacity of a linear network for the tracking task described here depends on whether the network inputs are raw grayscale natural images extracted from video, or sparse-code representations of those images. In Figure 5(a), a scatterplot of neighboring pixel values given by $v_1(x^s)$ and $v_2(x^s)$ is shown for $n$ samples ($s=1,..,n$) taken from grayscale natural images. 
\begin{figure}
\centering
\includegraphics[scale=0.7, bb=22 48 494 274, clip=true]{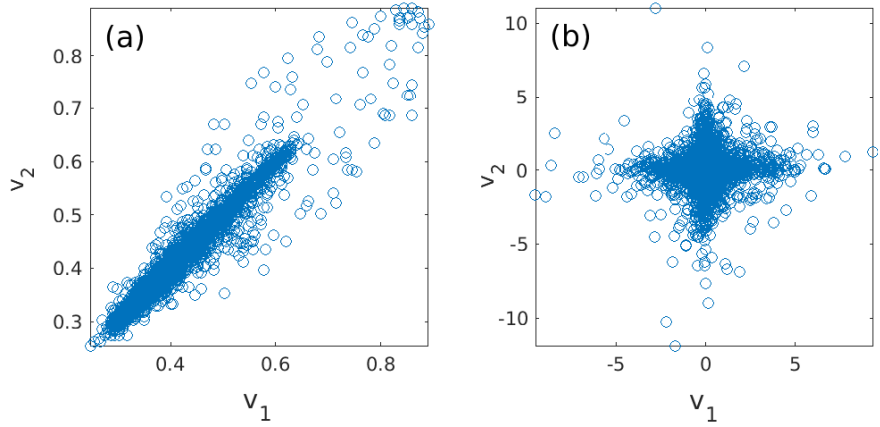}
\caption{Scatter plots of $v_1$ and $v_2$ samples taken from natural images (a), and their sparse code representations (b). In (a), neighboring pixels in a natural image show strongly correlated behaviour. In (b), neighboring Gabor coefficients in the sparse code show uncorrelated and symmetrical behaviour.}
\label{}
\end{figure} 
Neighboring pixels are highly correlated in natural images, so all samples fall into the first quadrant of the $(v_1,v_2)$ coordinate plane. This means the product $v_1(x^s)v_2(x^s)$ is always positive, and therefore $\sum_s v_1(x^s)v_2(x^s)>0$. The same conclusion is also true for correlated data with zero mean, as all samples then fall into quadrants one or three, and it is still the case that $v_1(x^s)v_2(x^s)> 0$. Therefore, the columns of $V$ must be non-orthogonal for grayscale natural image pixels, and $n_{max}<p$ is possible (linear dependence has not been proven, though in the examples presented here it is always the case). In Figure 5(b), a scatterplot of neighboring Gabor coefficient values given by $v_1(x^s)$ and $v_2(x^s)$ is shown for $n$ samples taken from the sparse code of these images. It can be seen that 2D Gabor function coefficients are uncorrelated and symmetrically distributed in the $(v_1,v_2)$ plane. Samples falling into the first or third quadrants contribute positive or zero terms to $\sum_s v_1(x^s)v_2(x^s)$, while samples falling into the second or fourth quadrants contribute negative or zero terms. As the samples are distributed symmetrically, the positive and negative contributions cancel due to symmetry, and $\sum_s v_1(x^s)v_2(x^s)\approx 0$. The columns of $V$ are therefore orthogonal for the sparse code, or close to orthogonal, and $n_{max}= p$. In summary, correlations between each pair of measurements $v_i(x^s)$ and $v_j(x^s)$ allow for $n_{max}<p$, while the sparse code decorrelates these measurements and approaches $n_{max}=p$.

To determine how large the effect of correlated measurements is in practice, memory capacity is now investigated quantitatively by finding the rank of $V$. The rank of $V$ is given by the number of non-zero singular values of $V$, which is approximated here by the number of singular values of $V\geq 0.1$. In Figure 6, curves for $n_{max}$ versus $p$ are generated by increasing $p$ (corresponding to the number of weights in the linear network) for each value of $n_{max}$, until achieving $n_{max}$ singular values of $V\geq 0.1$. 
\begin{figure}
\centering
\includegraphics[scale=0.7]{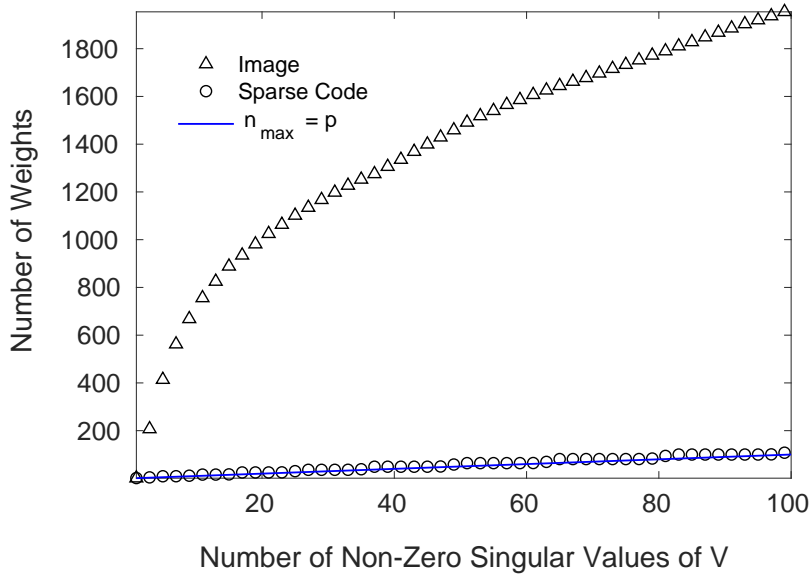}
\caption{Memory capacity of a linear network using network inputs given by natural images (triangles) or a sparse code (circles). The value of $p$, corresponding to the number of weights, must increase rapidly with increase in $n_{max}$, the number of non-zero singular values of the design matrix $V$. This happens more rapidly for a natural image input than for a sparse code, with the sparse code achieving the maximum memory capacity of a linear network, $n_{max}=p$.}
\label{}
\end{figure} 
It can be seen from Figure 6 that the sparse code achieves $n_{max}=p$, reaching the maximum memory capacity of a linear network. It can also be seen that using a natural image input leads to $n_{max}<p$, in agreement with the qualitative analysis presented above. Further, in Figure 6, it is also seen that using a natural image input requires up to 40 times the number of weights to store the same number of cost samples as using the sparse code. While this number is sensitive to the size of the cutoff chosen for the non-zero singular values, the qualitative behaviour of the figure is robust: correlated measurements have a noticeable effect on the memory capacity of a linear network. 

\subsection{Speed of learning}
The speed of learning for the linear network is given by the rate of convergence of the incremental gradient method in equation (\ref{grad}). The converged solution yields the cost-to-go approximation in equation (\ref{nn}) used for selecting optimal controls. A fast speed of learning implies a rapid rate of convergence, and requires less computational effort to arrive at a cost-to-go approximation. 

The rate of convergence of gradient methods often depends on the condition number of the Hessian matrix \citep{bert2,boyd}. This is given by the ratio of the largest and smallest eigenvalues of the Hessian matrix, and provides a measure of the eccentricity of level sets of expression (\ref{lsfit}) close to an optimal solution. Problems with a large condition number tend to have very elongated level sets, and gradient methods will often converge more slowly than for problems with a small condition number. In the present case, the Hessian matrix elements are proportional to the term $\sum_s v_i(x^s)v_j(x^s)$ which, as discussed in the previous section, depends only on the second-order statistics of a representation. From the qualitative analysis given in the previous section, the sparse code satisfies $\sum_s v_i(x^s)v_j(x^s)\approx 0$ for $i\neq j$; leading to an approximately diagonal Hessian with eigenvalues $\lambda_i\approx 2\sum_s v_i(x^s)^2$, which are proportional to the variance of $v_i$ for zero mean. Gabor coefficients of a sparse code tend to have similar variances (as seen in Figure 5(b)), and the condition number is therefore expected to be relatively small. When correlations are present in the network input, the variance along each principal axis of the input data may vary greatly. In Figure 5(a), the length of the elongated ellipse corresponds to the variance along one principal axis, and the width, to the variance along another principal axis. The diagonalized Hessian matrix will therefore have some eigenvalues that are much larger than others when strong correlations are present. For the natural image pixels shown in Figure 5(a), the condition number would therefore be expected to be relatively large.

In Figure 7, the rate of convergence of the incremental gradient method is given by a log-linear plot of $|V\hat{r}-\beta|$ averaged over 6 tracking tasks (giving the average error) versus number of iterations. 
\begin{figure}
\centering
\includegraphics[scale=0.7]{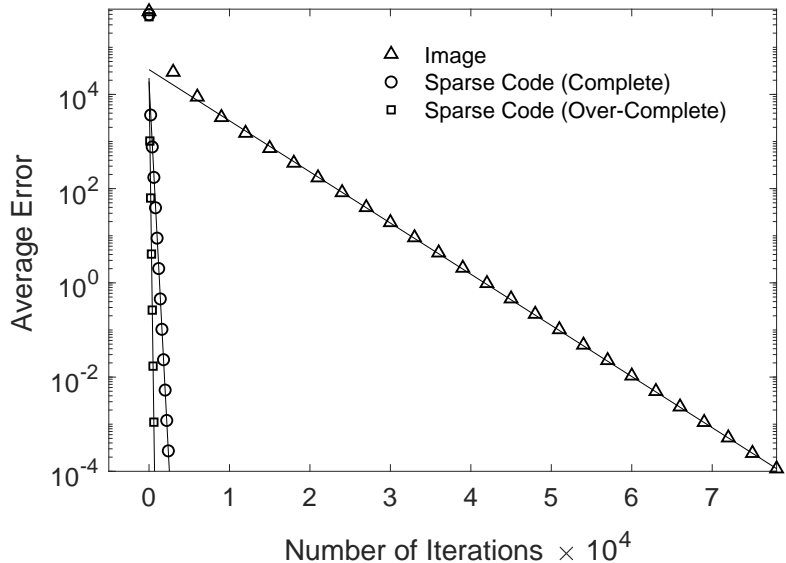}
\caption{Rate of convergence of the incremental gradient method for different image representations shown using a log-linear plot of average error versus number of iterations. The rate of convergence is at least an order of magnitude faster for the complete sparse code (circles) and the $\times 2.25$ over-complete sparse code (squares), than for the natural image input (triangles).}
\label{}
\end{figure}
Data is shown for both complete ($p=q$) and over-complete ($p=2.25q$) sparse codes, as well as for natural image inputs. Solid lines represent straight-line fits to data, indicating that average error decreases exponentially with the number of iterations. Assuming ``average error" $\propto e^{-t/\delta}$, time constants for the rates of convergence estimated from the data in Figure 7 give $\delta = 4000$ for natural image inputs, $\delta = 132$ for the complete sparse code, and $\delta = 36$ for the over-complete sparse code. Natural image inputs therefore require approximately 30 times the number of iterations of the complete sparse code to reach convergence. These results are in agreement with the qualitative discussion on condition numbers: poorly conditioned problems can take orders of magnitude longer to converge. The over-complete sparse code requires fewer iterations than the complete sparse code, however, the vector multiplication in equations (\ref{nn}) and (\ref{grad}) takes longer for an over-complete code as the vectors are longer.

\subsection{An over-complete sparse code for sequential learning}
Results for memory capacity and speed of learning only depend on the second-order statistics of image representations. On the other hand, a sparse code is distinguished from other decorrelated codes by (1) its higher-order statistics, and (2) its ability to form over-complete representations. Over-complete sparse codes turn out to be very useful for learning a large number of tasks sequentially, as will now be shown. An over-complete sparse code corresponds to $m>d$ in Equation (\ref{gabor}), so the number of 2D Gabor functions used in a representation is larger than the number of pixels in the underlying image. In this work, a $\times 2.25$ over-complete sparse code is used. In the language of neural networks this corresponds to doubling the number of inputs to a neural network or, more precisely, doubling the number of features extracted from a fixed-sized image input. These features live in a larger space than those of a complete sparse code, and yet remain approximately decorrelated by the sparse coding, as shown in Figure 8.
\begin{figure}
\centering
\includegraphics[scale=0.6]{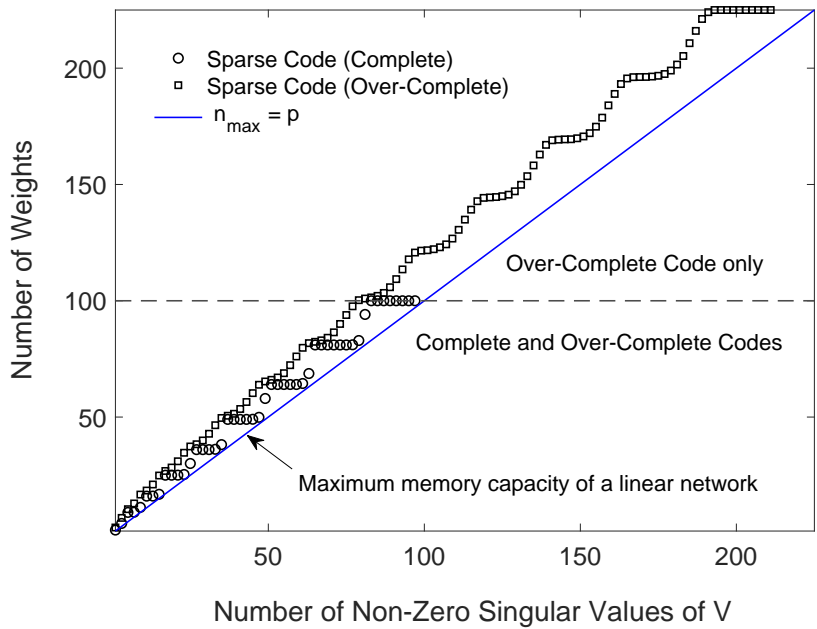}
\caption{Memory capacity of a linear network using an over-complete sparse code. Input images are $10\times 10$ pixels, giving up to $p=q=100$ weights for a complete sparse code (circles), and $p=2.25q=225$ weights for a $\times 2.25$ over-complete sparse code (squares). The over-complete sparse code closely follows the maximum memory capacity of a linear network (blue line) as the number of weights are increased, and complete codes become no longer possible (above dashed line). A linear network using this over-complete sparse code can efficiently store over 200 cost values for a $10\times 10$ input image. Non-zero singular values of $V$ are singular values of $V\geq 0.1$, as in Figure 6.}
\label{}
\end{figure}
The memory capacity of a linear network therefore at least doubles in size using a $\times 2.25$ over-complete sparse code. An over-complete code with correlated features would require a much larger degree of over-completeness, and correspondingly many more network weights, in order to achieve a similar increase in memory capacity.

Neuro-dynamic programming allows multiple tracking tasks to be learned sequentially when a gradient method is used to update network weights. Optimal control of multiple tracking tasks is then possible, and new tracking tasks can be learned as soon as the data become available. However, as each new task is learned and the network is updated, it becomes more likely that previously trained tasks are ``forgotten". This likelihood increases as the representations of different tasks begin to overlap and interfere (as in the case of two very similar tasks with very different costs), or when memory capacity is exceeded. In the first case, overlap between a new task and a previously trained task would result in un-learning the earlier task during the process of learning the new task, even when below memory capacity. In the neural network community this is known as ``catastrophic forgetting" \citep{kirk}. One possible solution is to re-train on the entire past sequence of tracking tasks. This can be done either by joint optimization of all tracking tasks as a single batch, or by repeating the sequence of tracking tasks many times during sequential learning. Both methods require access to past tracking tasks, which is often not available; and both methods must re-process past tracking tasks each time a new task is learned, which is computationally inefficient. Instead, a new approach to sequential multitask learning is proposed that circumvents the need for past data and additional processing.

Catastrophic forgetting is now investigated for the case of multiple tracking tasks. In Figure 9, performance on four tracking tasks is shown using either natural image pixels as input, a decorrelated code that is not sparse, or a sparse code. 
\begin{figure}
\centering
\includegraphics[scale=0.6]{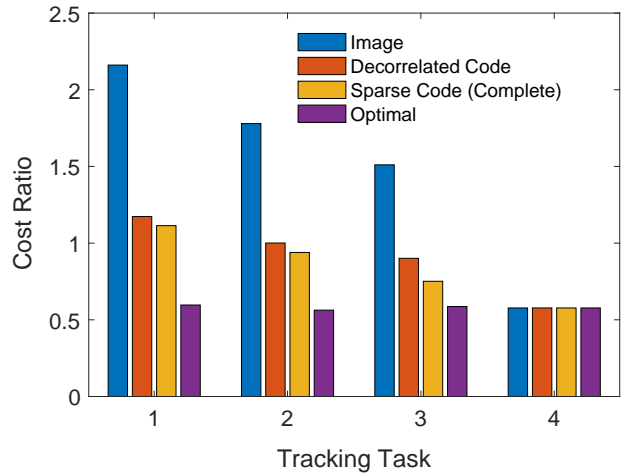}
\caption{Sequential learning of multiple tracking tasks leading to catastrophic forgetting. Tracking tasks were learned in ascending order of task number, and cost ratio is the ratio of neuro-DP cost to greedy cost. Tracking performance is optimal for tracking task 4 (the most recently learned task), with performance declining for earlier tasks. Data represent an average of 200 random combinations of 4 tracking tasks taken from a total of 8.}
\label{}
\end{figure}
The decorrelated code that is not sparse is found using filter-based decorrelation, as described in \cite{olshausen96,olshausen97}, and \cite{hyvarinenbook}. Filtering with the discrete Fourier transform means a filter-based decorrelated code has the same dimensionality (i.e., number of pixels) as the underlying image, and therefore cannot be made into an over-complete code. The same is true for alternative decorrelation methods such as patch-based decorrelation using PCA \citep{hyvarinenbook}. By contrast, a sparse code is decorrelated and can be made over-complete due to the fact it is a solution to a least-squares problem.

Each tracking task in Figure 9 was learned sequentially by iterating the neuro-DP equations until convergence, then transferring the weight values $r_k$ to the next task in the sequence, and repeating. Tracking tasks were learned in ascending order of task number; i.e., task 1 first, etc. The performance measure is given by ``cost ratio", which is the total cost of the neuro-DP solution divided by the total cost of the greedy solution. When ``cost ratio" is less than one this means neuro-DP has found a lower cost solution than the greedy solution: otherwise, the solution is either of equal cost, or higher cost. The optimal solution found by applying exact DP to each tracking task without sequential learning is also shown for comparison. It is seen that tracking performance is only optimal for tracking task four: the most recently learned task, with performance declining rapidly for earlier tasks due to catastrophic forgetting. Tracking performance is seen to decline most rapidly for network inputs given by natural images, where it is worse than greedy for all tracking tasks other than tracking task four. Tracking performance for the decorrelated code, and the complete sparse code, is substantially better. The reason is that natural images, sparse codes, and decorrelated codes are all distributed representations, so different tracking tasks are likely to overlap to some extent. Overlap of different tasks in a representation can sometimes lead to desirable generalization if the state-cost pairs have similar values for the different tasks, or more often, to unwanted interference if they do not. On average, the sparse code and the decorrelated code have less overlap between different tasks than do natural images (see next section for details). This reduces the chance of unwanted interference  during sequential learning, and leads to slightly  better tracking performance, as seen in Figure 9. Filter-based decorrelation will not be considered any further in this work. 

To avoid catastrophic forgetting the approach taken here is to divide each representation into equal non-overlapping regions called partitions, and then solve each tracking task on a unique partition. These partitioned representations exclude the possibility of interference between different tracking tasks, allowing for generalization within tasks but not between different tasks. Tasks that do generalize well without detrimental interference can be placed on the same partition. The vector $v_k(x_k^s)$ then becomes $v_k(x_k^s, i)$: where $i$ is the partition index, and $v_k(x_k^s, i)$ is a vector of all zeros except for a small region of sparse code or natural image forming the partition. This approach works by limiting the updates of $r_k$ in equation (\ref{grad}) to the region of non-zero support in $v_k(x_k^s, i)$. A simple memory capacity argument can now be used to establish the maximum number of useful partitions for each type of representation. Using an estimate of 30 states per stage of a tracking task from the last panel of Figure 3 (every state is used as a training sample in this case) leads to $n=30$, giving $p=1200$ from Figure 6 when the network input is a natural image. For an image region of size $110\times 110$ pixels, $q=12100$, and memory capacity for a natural image partition would be achieved after $q/1200=10.1$ partitions. This means up to ten tracking tasks could be learned sequentially and stored in the network at any one time, returning optimal tracking performance for any of these tasks at a later point in time. For a complete sparse code, $n=p$ gives $p=30$, and memory capacity would be reached after $q/30=403$ partitions. For a $\times 2.25$ over-complete sparse code, memory capacity would be reached after $403\times 2.25=907$ partitions. This simple argument shows the large improvements possible from decorrelation and over-completeness.

In practice, the maximum number of partitions is difficult to achieve in numerical experiments with equation (\ref{grad}) because sensible values for the learning rate become difficult to determine when approaching memory capacity. This is the case in Figures 10 and 11; where the results are easy to achieve, but fall short of reaching the maximum number of partitions. 
\begin{figure}
\centering
\includegraphics[scale=0.6]{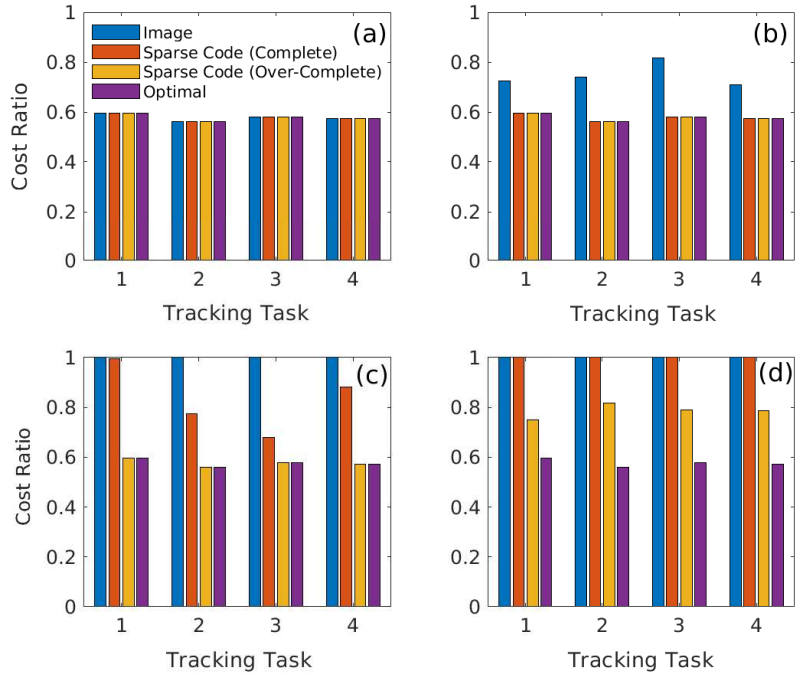}
\caption{Sequential learning of multiple tracking tasks using partitioned representations. In (a), 4 partitions lead to optimal tracking performance on 4 tracking tasks for all representations. In (b), 32 partitions (of which only 4 are used) lead to suboptimal tracking performance for natural image inputs. In (c), 256 partitions lead to suboptimal tracking performance for the complete sparse code. In (d), 512 partitions lead to suboptimal tracking performance for the over-complete sparse code.}
\label{}
\end{figure}
Nevertheless, the general principle  can easily be demonstrated. Figure 10 is a repeat of the calculation done in Figure 9 using a varying number of partitions. In Figure 10(a), four partitions give optimal performance for all representations on four tracking tasks. In Figure 10(b), each representation is divided into 32 partitions and four of these are used to solve four tracking tasks. It is seen that tracking is no longer optimal for natural image partitions. In Figure 10(c), each representation is divided into 256 partitions and tracking becomes suboptimal for the complete sparse code partitions. Finally, in Figure 10(d), each representation is divided into 512 partitions and tracking becomes suboptimal for the over-complete sparse code partitions as well.

The results in Figure 10 are summarized in Figure 11.
\begin{figure}
\centering
\includegraphics[scale=0.7]{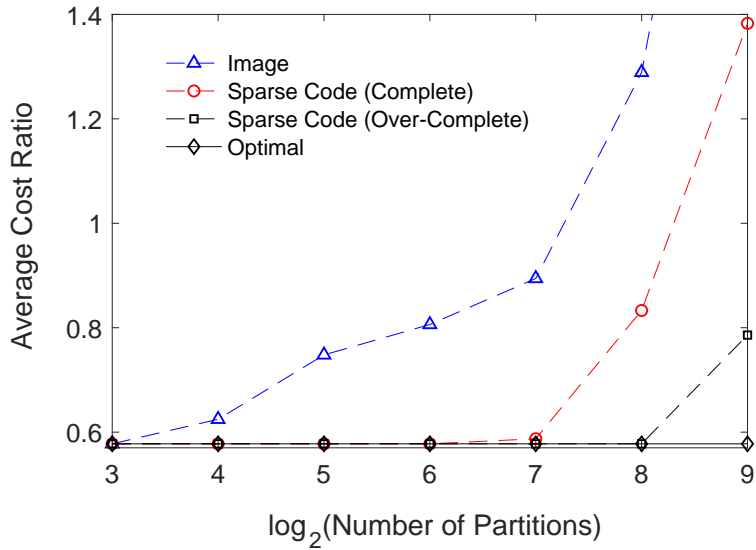}
\caption{Effect of increasing the number of partitions from 8 to 512 on the average performance of partitioned representations for multiple tracking tasks. Natural image partitions (triangles) remain optimal at $2^3=8$ partitions, but are suboptimal by $2^4=16$. Complete sparse code partitions (circles) remain optimal at $2^7=128$ partitions, and $ \times 2.25$ over-complete sparse code partitions (squares) remain optimal at $2^8=256$ partitions.}
\label{}
\end{figure}
Cost ratio is averaged over the four tracking tasks in Figure 10 to give ``average cost ratio" in Figure 11. The horizontal axis in Figure 11 is log base 2 of the number of partitions; starting at $2^3=8$, then doubling in number until reaching $2^9=512$. The blue curve shows that natural images are optimal at $2^3=8$ partitions, and are suboptimal by $2^4=16$ partitions. The red and green curves show that the complete sparse code is still optimal at $2^7=128$ partitions, and the over-complete sparse code, at $2^8=256$ partitions. These results fall short of the maximum number of partitions for the reason previously discussed. However, memory capacity could be further increased in the case of the over-complete sparse code simply by increasing its degree of over-completeness. 

Dividing the sparse code into partitions has clear advantages for avoiding catastrophic forgetting in sequential multitask learning. The over-complete sparse code offers an additional advantage of being able to further increase memory capacity through over-completeness. In this work, a $\times 2.25$ over-complete sparse code was used. However, the degree of over-completeness can be increased well beyond this range \citep{sommer,olshausen13b}, potentially allowing many more tasks to be learned sequentially. Further, each partition can be accessed with a key that depends on some unique aspect of a tracking task. For example, a key could be implemented as a hash function $f$ of the initial location of the target (i.e., $k=1,s=\mathrm{Target}$): giving $i=f(x_{1}^{\mathrm{Target}})$, or of the image region associated with the initial location of the target: giving $i=f(v(x_{1}^{\mathrm{Target}}))$. It is also important to remember that each partition yields a function approximator for the cost-to-go. Compared with using a table of cost-to-go values as in exact DP, this approach has the advantage of scaling up to arbitrarily large problems where the number of states can be much larger than either the number of network weights in any partition, or the number of training samples available. In other words, within each partition, the function approximator can generalize over states at the expense of possibly becoming suboptimal, while a method based on tables cannot.

\subsection{Sparse code representational capacity, redundancy, and fault tolerance}
A distributed representation makes combinatorial use of its primitive elements to represent the different states of a system. An example is the binary representation of integers, where each primitive element is a single binary digit with 2 possible levels given by 0 or 1, and $N$ binary digits combine together to represent $2^N$ possible integers. When all integers occur with equal probability, the average information content of each binary digit is 1 bit (as given by the entropy: $H(X)=\sum_{x\in \mathrm{Im} X}p(x)\log_21/p(x)$), and the number of ``typical" states \citep{mackay} of $N$ independent binary digits becomes $2^{NH}=2^N$. In this case the number of ``typical" states equals the total number of states and no compression is possible. 

In the present case, grayscale natural images and complete sparse codes are both distributed representations. Firstly, consider an 8-bit grayscale image with i.i.d. pixel values chosen uniformly at random. Each pixel has an entropy of 8 bits, and the number of ``typical" random images with $N$ pixels is $2^{NH}=256^N$. Now consider a grayscale natural image with a histogram of pixel values as shown in Figure 12(a).
\begin{figure}
\centering
\includegraphics[scale=0.7]{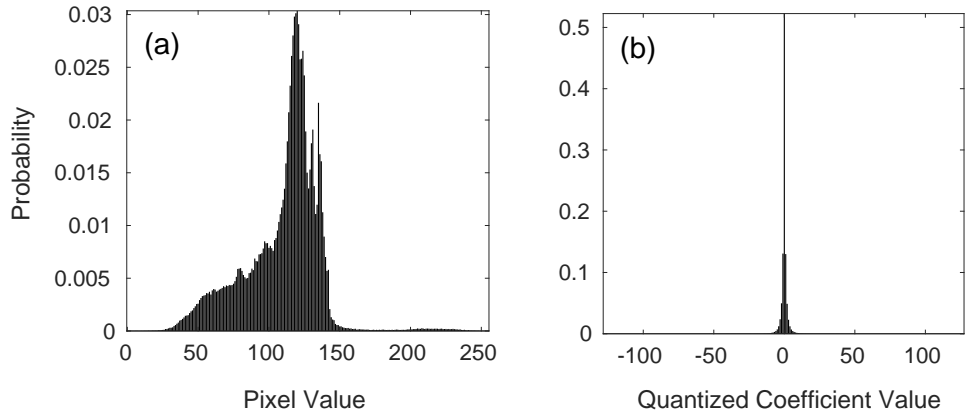}
\caption{In (a), a histogram of pixel values taken from a grayscale natural image. In (b), a histogram of quantized Gabor coefficient values taken from the (complete) sparse code of the image used in (a).}
\label{}
\end{figure} 
As all pixel values are not used with uniform frequency, each pixel has a slightly lower entropy of 6.5 bits. Applying the formula $2^{NH}$ to find the number of ``typical" grayscale natural images with $N$ pixels gives $2^{NH}\approx 91^N$ for sufficiently large $N$. However, as seen in Figure 5(a), these pixel values are not i.i.d.~because they are strongly correlated, so this formula is not valid and the real number must be less than this value.
 
Applying a QP solver to solve equation (\ref{gabor}) leads to a set of double-precision floating-point numbers for the Gabor coefficients $a({r}^{\prime})$. To compare entropies with grayscale images, these coefficients are quantized to 8-bit integers by applying the method of uniform scalar quantization \citep{gal}. The histogram in Figure 12(b) was constructed by binning all coefficient values over the range -128.5 to 127.5 into 256 bins, leading to an entropy of 2.5 bits per quantized Gabor coefficient. This construction slightly overestimates the entropy as coefficient values in this example run from -137.2 to 140.8: sparse distributions are heavy tailed and strongly peaked about zero. Entropy estimates for decorrelated codes that are not sparse codes turn out to be quite similar to those for sparse codes: see, for example, PCA versus ICA in \cite{eichhorn}. Although quantization of real-valued Gabor coefficients results in some distortion of reconstructed images, as does the least squares approximation in (\ref{gabor}), it also leads to an estimate of representational capacity. Since Gabor coefficients are closer to being i.i.d.~as seen by the reduced dependency exhibited in Figure 5(b), the number of ``typical" sparse codes of $N$ quantized Gabor coefficients can be approximated by $2^{NH}=2^{2.5N}\approx 6^N$ for sufficiently large $N$. This is a better estimate of the number of ``typical" grayscale natural images, however, some statistical dependency will still be present (see \cite{hyvarinenbook}). An image model that removes all statistical dependencies in its coefficients and can reconstruct images with zero distortion would give a precise value. Unfortunately, no such model is currently known. 

It is now helpful to distinguish between statistical redundancy, due to the presence of statistical dependency in a representation, and redundancy in a representation's capacity to represent different states. The key point is that the number of different images that can be represented using either grayscale natural images or a complete sparse code grows exponentially with $N$, so both representations are highly redundant: more images can be represented than would ever be experienced (i.e., for a $10\times 10$ image, $N=100$, and $6^N=10^{77}$). This shows that a sparse code has low statistical redundancy, but high redundancy in its representational capacity.

Another form of redundancy in a sparse code is seen in Figure 12(b): the value of a Gabor coefficient is more likely to be zero than any other value. Statistical dependency has been reduced but redundancy is now present in the coefficient frequencies. This form of redundancy can be removed using the technique of arithmetic coding \citep{mackay}, allowing Gabor coefficient values to be written to a binary file whose size is within $NH+2$ bits. However, this form of redundancy turns out to be useful for encoding multiple tasks using a linear network, since the chance of overlap between different tasks is reduced. In Figure 12(b), 20\% of the Gabor coefficients have values in the range $(-0.1, 0.1)$ before quantization. In Figure 12(a), only 0.05\% of the grayscale image pixels have values in the range $(0, 0.1)$ after converting to double precision floating-point numbers between 0 and 1 for use in neuro-DP. The larger relative number of Gabor coefficients close to zero decreases the chance that two or more ``sparse" representations will overlap in a sparse code, compared with using grayscale natural images. This results in less unwanted interference when learning multiple tasks sequentially. It can also be interpreted from Figure 9 that there is less interference between tasks in the sparse code and the decorrelated code than there is in grayscale natural images. At the other extreme, \emph{local codes} (codes that use $N$ elements to represent $N$ states) have no overlap and do not suffer from unwanted interference. However, the representational capacity of a local code is too limited for most purposes.

Distributed representations are tolerant to faults and can continue to function even when a network is badly damaged. In the present case, randomly corrupting some of the weights of the linear network will reduce its tracking performance but not completely destroy its tracking ability. The fault tolerance of a linear network is directly related to the redundancy in its weights. From the previous discussion of memory capacity, a grayscale natural image (which is an example of a \emph{dense code}, since it is not sparse) requires many times more network weights than a complete sparse code to store the same number of target values in a linear network. This redundancy means a network trained on natural image inputs will have a higher fault tolerance than one trained on sparse codes: changing one of the weights of the network is less likely to have an effect on the output of the network.

\section{Conclusion}\label{conclusion}
In this investigation, an optimal control approach to coding was taken, connecting representations of sensory data to the computational problem-solving tasks at hand. The central finding was that the combination of decorrelation and over-completeness gives a sparse code a computational advantage over other codes for optimal control tasks with correlated feature inputs. The main conclusions of this work are the following. 

A complete sparse code was shown to maximise the memory capacity of a linear network by transforming the design matrix of the least-squares problem to be one of full rank. Neighboring pixels in natural images represent strongly correlated measurements of natural scenes, leading to a design matrix that is below full rank and therefore unable to store the maximum number of target values. A sparse code was shown to decorrelate these measurements and form an orthogonal basis for the design matrix.

A sparse code was also shown to increase the speed of learning the network weights by conditioning the Hessian matrix of the associated least-squares problem. Large covariances in natural images lead to a poorly conditioned Hessian matrix describing highly elongated level sets, and decreasing the rate of convergence to the optimal solution for the network weights. A sparse code increases this rate of convergence by virtue of a well-conditioned Hessian matrix. 

The central result of this work was that an over-complete sparse code was found to increase the memory capacity of a linear network beyond that possible for a complete code with the same-sized input, while still providing a basis for the design matrix that is close to orthogonal (i.e., the resulting matrix is close to full rank). This was demonstrated for a $\times 2.25$ over-complete sparse code, where the memory capacity at least doubles compared with a complete sparse code using the same input. It was also shown how to use an over-complete sparse code to sequentially learn a potentially large number of optimal control tasks while avoiding catastrophic forgetting.

An investigation of the representational capacity and redundancy of a sparse code showed it to be a highly redundant representation capable of representing more images than would ever be experienced, while also being biased towards values close to zero, thereby decreasing the chance that representations of two or more optimal control tasks will overlap and interfere during sequential learning.

\subsection{Future work}
In practical applications, sparse codes could be used for solving Markov decision processes or optimal control tasks with an infinite horizon and a large number of states. A key question then becomes how over-complete can a sparse code be made while still providing a useful representation as a function approximator? It is clear that highly over-complete sparse codes are possible \citep{sommer,olshausen13b}. It can be seen in Figure 8 that going from a complete sparse code, to a $\times 2.25$ over-complete sparse code leads to a slight loss of memory capacity. This might be rectified using a highly over-complete sparse code. Increasing the sparsity of a code may also help further reduce interference between tasks during sequential learning.  

 It is clear the neuro-dynamic programming model proposed here is not useful as an online tracking algorithm in its current form. Nonetheless, a straightforward extension allows it to be used for online re-planning. Firstly, it should be noted that a large class of popular online object tracking algorithms would not be capable of finding optimal solutions to the type of tracking tasks considered here. For example, kernal-based object tracking employs a mean-shift procedure to locate the local maxima of a function measuring similarity between target and candidate probability distribution functions \citep{com}. However, this is a greedy approach, and in the case of restricted controls will generally lead to suboptimal solutions. To find optimal solutions the neuro-dynamic programming model requires each tracking task to be processed off-line. These pre-trained tasks can then be modified online using the \emph{rollout algorithm} \citep{bert} by making use of the learned function approximator from neuro-dynamic programming to generate a base policy, which is then improved online in a one-step lookahead by rollout. The improved policy is a sequence of controls that have been updated for the current task. This does require a model for the updated parts of the target trajectory in order to simulate roll out of the base policy. Provided a pre-trained task contributes enough information about the non-greedy stages of the online task (for example, stages where the restricted set of controls may lead to a bottleneck when following a target trajectory), this extension should be better suited to finding an optimal tracking solution online when suboptimal solutions are also present.

Partitioning representations to avoid catastrophic forgetting might be useful for competitive reinforcement learning \citep{mckenzie}. Distributed representations are strongly susceptible to overlap between different tasks, and if these tasks do not generalize well this will lead to unwanted interference during sequential learning. Using over-complete sparse codes, and specializing different regions of a network to different tasks, should help eliminate catastrophic forgetting while making most efficient use of the resources available. Similar tasks that generalize well can share network regions, while tasks that compete strongly during learning would make use of seperate network regions. 

\subsection*{Acknowledgements}
I thank John Daugman and Bruno Olshausen for useful discussions related to this work. I thank Steven Wiederman for supplying the video used in this study. Part of this work was completed on sabbatical at the Redwood Center for Theoretical Neuroscience, University of California, Berkeley.

\section*{Appendix}
\subsection*{A1: A one-dimensional target tracking problem}
Consider a one-dimensional target tracking problem with restricted controls over a finite horizon. At time $k$, the state $x_k$, and the target location $w_k$, only take values given by the non-negative integers $0,1,2,...$, and the controls are restricted to be either $u_k=0$ or $u_k=1$. The DP algorithm corresponding to equation (\ref{dp}) is given by $J_N(x_N)=0$, and
\begin{align}
J_{k}(x_{k})&=\underset{u_k\in \{0,1\}}{\operatorname{min}}\mathbb{E}\left[|x_k-w_k|+J_{k+1}(x_{k}+u_k)\right],\nonumber\\
&=\text{min}\left[J_{k+1}(x_{k}),J_{k+1}(x_{k}+1)\right]+\mathbb{E}\left[|x_k-w_k|\right].\label{simpledp}
\end{align}

In the deterministic version of this problem, a simple way to include suboptimal greedy solutions is to consider the horizon to be at $N=4$ and set the target location to the following values:
\begin{equation*}
\begin{aligned}
& w_0=0, && w_1=0, &&&w_2=2, &&&&w_3=3.
\end{aligned}
\end{equation*}
The total cost to be minimized is then given by
\begin{align*}
\sum_{k=0}^{3}|x_k-w_k|&=|x_0|+|x_1|+|x_2-2|+|x_3-3|.
\end{align*}   
Applying backward iteration to the DP algorithm (\ref{simpledp}) leads to the following equations:
\begin{align*}
&J_4(x_4) = 0,\\
&J_3(x_3) = |x_3-w_3|=|x_3-3|,\\
& J_2(x_2)=\text{min}\left[J_{3}(x_{2}),J_{3}(x_{2}+1)\right]+|x_2-2|,\\
& J_1(x_1)=\text{min}\left[J_{2}(x_{1}),J_{2}(x_{1}+1)\right]+|x_1|,\\
& J_0(x_0)=\text{min}\left[J_{1}(x_{0}),J_{1}(x_{0}+1)\right]+|x_0|.
\end{align*}
We can now fill out the first row of the table of cost values:
\begin{equation*}
\begin{aligned}
&J_3(0) = 3, &&  J_3(1) = 2, &&& J_3(2) = 1, &&&& J_3(3)=0.
\end{aligned}
\end{equation*}
These cost values decrease with increase in state, so that in the second row of the table: $\text{min}\left[J_{3}(x_{2}),J_{3}(x_{2}+1)\right]=J_{3}(x_{2}+1)$, and $u_2^*=1$. The second row of the table now becomes
\begin{equation*}
\begin{aligned}
&J_2(0) = 4, &&  J_2(1) = 2, &&& J_2(2)= 0, &&&&\text{with } u_2^*=1.
\end{aligned}
\end{equation*}
Following this general pattern for the other rows gives,
\begin{equation*}
\begin{aligned}
&J_1(0) = 2, &&  J_1(1) = 1, &&& \text{with } u_1^*=1,\\
&J_0(0) = 1, && \text{with } u_0^*=1.
\end{aligned}
\end{equation*}
The total cost of the DP solution is therefore $J_0(0) =1$, and the optimal controls $u^*=(1,1,1)$ lead to the optimal tracking trajectory $x^*=(0, 1, 2, 3)$.

Let's compare this result with the greedy policy. A greedy policy would return the control leading to the lowest cost at the next time period without regard for future times. This policy corresponds to minimizing only the expectation term in equation (\ref{simpledp}). A greedy tracker would therefore select the controls $u=(0, 1, 1)$ leading to the tracking trajectory $x=(0, 0, 1, 2)$, and a total cost of 2. The greedy solution is therefore suboptimal, while the DP solution is optimal. The nature of the optimal solution is that it requires the optimal controls $u^*=(1,1,1)$ to prevent the tracker being outrun by the target. The greedy tracker, by contrast, takes the controls $u=(0,1,1)$, allowing the target to get ahead of the tracker, which is never able to catch up. This optimality is a function of the horizon. Putting the horizon at $N=3$, the greedy and DP solutions are both optimal. For $N>3$, the greedy solution has a cost of $N-3$ greater than the DP solution if the target continues moving towards the horizon.

In the stochastic version of this problem, suboptimal greedy solutions can be included by putting the horizon at $N=3$, and allowing the target location to take random values from a non-stationary probability distribution given by
\begin{equation*}
\begin{aligned}
& p(w_0=0)=1, && p(w_0=1)=0, &&& p(w_0=2)=0,\\ 
& p(w_1=0)=p_1, && p(w_1=1)=1-p_1, &&& p(w_1=2)=0,\\ 
& p(w_2=0)=0, && p(w_2=1)=1-p_2, &&& p(w_2=2)=p_2.
\end{aligned}
\end{equation*}
Importantly, we assume $p_k>1-p_k$ at each time $k$, so that $p_k>1/2$. The total cost to be minimized is now
\begin{align*}
\mathbb{E}\left[\sum_{k=0}^{2}|x_k-w_k|\right]&=|x_0|+p_1|x_1|+(1-p_1)|x_1-1|\nonumber\\
&+(1-p_2)|x_2-1|+p_2|x_2-2|.
\end{align*}   
Applying backward iteration to the DP algorithm (\ref{simpledp}) leads to the following equations:
\begin{align*}
&J_3(x_3) = 0,\\
&J_2(x_2) = \mathbb{E}\left[|x_2-w_2|\right]=(1-p_2)|x_2-1|+p_2|x_2-2|,\\
& J_1(x_1)=\text{min}\left[J_{2}(x_{1}),J_{2}(x_{1}+1)\right]+(1-p_1)|x_1-1|+p_1|x_1|,\\
& J_0(x_0)=\text{min}\left[J_{1}(x_{0}),J_{1}(x_{0}+1)\right]+|x_0|.
\end{align*}
We can now fill out the first row of the table of cost values:
\begin{equation*}
\begin{aligned}
&J_2(0) = 1+p_2, &&  J_2(1) = p_2, &&& J_2(2) = 1-p_2.
\end{aligned}
\end{equation*}
Following our assumption $p_2>1-p_2$, we see the cost values decrease with increase in state, so that in the second row of the table: $\text{min}\left[J_{2}(x_{1}),J_{2}(x_{1}+1)\right]=J_{2}(x_{1}+1)$ and $u_1^*=1$. The second row of the table now becomes
\begin{equation*}
\begin{aligned}
&J_1(0) = 1-(p_1-p_2), &&  J_1(1) = 1-(p_2-p_1), &&& \text{with } u_1^*=1.
\end{aligned}
\end{equation*}
The final row of the table depends on the relative values of $p_1$ and $p_2$. For $p_2>p_1$, 
\begin{equation*}
\begin{aligned}
&J_0(0) = 1-(p_2-p_1), &&\text{with } u_0^*=1.
\end{aligned}
\end{equation*}
While for $p_2<p_1$, 
\begin{equation*}
\begin{aligned}
&J_0(0) = 1-(p_1-p_2), &&\text{with } u_0^*=0.
\end{aligned}
\end{equation*}
To gain more insight into this solution let's compare it with the greedy policy. A greedy tracker would take controls $u_0=0$ (we assumed $p_1>1-p_1$), and $u_1=1$ ($u_1=2$ is not a control), corresponding to the trajectory $x=(0, 0,1)$. The corresponding cost is: $1-(p_1-p_2)$. Therefore, the cost of the greedy solution matches the cost of the DP solution $J_0(0) = 1-(p_1-p_2)$ when $p_2<p_1$. However, when $p_2>p_1$, the cost of the DP solution is $J_0(0) = 1-(p_2-p_1)$. This is $2(p_2-p_1)$ less than the cost of the greedy solution: making the greedy solution suboptimal, and the DP solution optimal.

\subsection*{A2: Sparse coding as a quadratic program}
Adding an $\ell_{1}$ regularization term to the least-squares approximation in equation (\ref{gabor}) gives 
\begin{equation}
\begin{aligned}
&\underset{a}{\operatorname{min}} & & \|Ga - I \|_{2}^{2}+\lambda\|a\|_{1}.\label{l1}
\end{aligned}
\end{equation}
This is an unconstrained optimization with a non-differentiable objective. However, as discussed in \cite{tib}, it can be transformed to a constrained optimization with a differentiable objective and solved as a quadratic program (QP). This is done by expressing the variable $a$ as the difference of two nonnegative variables $a^+$ and $a^-$, such that: $a=a^+-a^-$,  $\|a\|_{1}=a^++a^-$, and $a^+,a^-\geq 0$ (for $a\in\mathbb{R}$ it is easy to confirm the unique solution to these equations gives: $|a|=a$ for $a\geq 0$, and $|a|=-a$ for $a< 0$, as expected). Eq.(\ref{l1}) then becomes
\begin{equation}
\begin{aligned}
&\text{minimize} & & \|G(a^+-a^-) - I \|_{2}^{2}+\lambda (a^++a^-), \label{l1trans}\\
&\text{subject to} && a^+\geq 0, a^-\geq 0,
\end{aligned}
\end{equation}
which can be written in the form of a QP: 
\begin{equation*}
\begin{aligned}
&\text{minimize} & & \frac{1}{2}y^TPy+q^Ty \label{qp}\\
&\text{subject to} && y\geq l,
\end{aligned}
\end{equation*}
where $y=(a^+, a^-)$, $l=(0, 0)$, $q=(\lambda - 2G^TI, \lambda+2G^TI)$, and
\begin{equation*}
P=
\begin{bmatrix*}[l]
\phantom{-}2G^TG& -2G^TG\\
-2G^TG& \phantom{-}2G^TG
\end{bmatrix*}.
\end{equation*}
However, the number of variables to solve for has now doubled from $a$, to $a^+$ and $a^-$.


\begin{thebibliography}{100}
\providecommand{\natexlab}[1]{#1}
\expandafter\ifx\csname urlstyle\endcsname\relax
  \providecommand{\doi}[1]{doi:\discretionary{}{}{}#1}\else
  \providecommand{\doi}{doi:\discretionary{}{}{}\begingroup
  \urlstyle{rm}\Url}\fi



\bibitem[{Barlow(1961)}]{barlow}
Barlow, H.~B. (1961).
\newblock Possible principles underlying the transformation of sensory messages. 
\newblock \emph{Sensory Communication}, 217--234.

\bibitem[{Bertsekas(2017)}]{bert}
Bertsekas, D.~P. (2017).
\newblock \emph{Dynamic programming and optimal control vol 1, 4th ed}.
\newblock Athena Scientific.

\bibitem[{Bertsekas \& Tsitsiklis(1996)}]{bert2}
Bertsekas, D.~P., \& Tsitsiklis, J.~N. (1996).
\newblock \emph{Neuro-dynamic programming}.
\newblock Athena Scientific.

\bibitem[{Boyd \& Vandenberghe(2004)}]{boyd}
Boyd, S., \& Vandenberghe, L. (2004).
\newblock \emph{Convex Optimization}.
\newblock Cambridge University Press.

\bibitem[{Comaniciu et al.(2003)}]{com}
Comaniciu, D., Ramesh, V., \& Meer, P. (2003).
\newblock Kernel-based object tracking.
\newblock \emph{IEEE Trans. Pattern. Anal. and Mach. Intell.}, \emph{25}, 564--577.

\bibitem[{Daugman(1985)}]{daugman85}
Daugman, J.~G. (1985).
\newblock Uncertainty relation for resolution in space, spatial frequency, and orientation optimized by two-dimensional visual cortical filters.
\newblock \emph{J.~Opt.~Soc.~Am.~A}, \emph{2}, 1160 -- 1169.

\bibitem[{Daugman(1988)}]{daugman88}
Daugman, J.~G. (1988).
\newblock Complete discrete 2-D Gabor transforms by neural networks for image analysis and compression.
\newblock \emph{IEEE Trans.~Acoustics, Speech, Sig.~Proc.}, \emph{36}, 1169 -- 1179.

\bibitem[{Daugman(1989)}]{daugman89}
Daugman, J.~G. (1989).
\newblock Entropy reduction and decorrelation in visual coding by oriented neural receptive fields.
\newblock \emph{IEEE Trans.~Biomed.~Eng.}, \emph{36}, 107 -- 114.

\bibitem[{Eichhorn et.~al.(2009)}]{eichhorn}
Eichhorn, J., Sinz, F., Bethge, M. (2009).
\newblock Natural Image Coding in V1: How Much Use Is Orientation Selectivity?
\newblock \emph{PLOS Comp.~Biol.}, \emph{5}, e1000336.

\bibitem[{Field(1994)}]{field}
Field, D.~J. (1994).
\newblock What is the goal of sensory coding?
\newblock \emph{Neural Comp.}, \emph{6}, 559 -- 601.

\bibitem[{F{\"o}ldi{\'a}k}(2002)]{fold}
F{\"o}ldi{\'a}k, P. (2002).
\newblock Sparse coding in the primate cortex.
\newblock \emph{The Handbook of Brain Theory and Neural Networks . 2nd edn}.
\newblock M A Arbib (ed.), MIT Press, 1064--1068.


\bibitem[{Gallager} (2008)]{gal}
Gallager, R.~G. (2008).
\newblock \emph{Principles of digital communication}.
\newblock Cambridge University Press.

\bibitem[{Hertz et.~al.(1991)}]{hertz}
Hertz, J., Krogh, A., Palmer, R.~G. (1991).
\newblock \emph{Introduction to the theory of neural computation}.
\newblock Addison-Wesley Publishing Company.

\bibitem[{Hyv\"arinen et.~al.(2009)}]{hyvarinenbook}
Hyv\"arinen, A., Hurri, J., \& Hoyer, P.~O. (2009).
\newblock \emph{Natural Image Statistics}, Springer-Verlag, London.

\bibitem[{Jones and Palmer(1987)}]{jones}
Jones, J.~P. \& Palmer, L.~A. (1987).
\newblock An evaluation of the two-dimensional Gabor filter model of simple receptive fields in cat striate cortex.
\newblock \emph{J.~Neurophys.}, \emph{58}, 1233 -- 1258.

\bibitem[{Kirkpatrick et.~al.(2017)}]{kirk}
Kirkpatrick, J., et.~al. (2017).
\newblock Overcoming catastrophic forgetting in neural networks.
\newblock \emph{PNAS}, \emph{114}, 3521 -- 3526.


\bibitem[{Loxley(2017)}]{loxley}
Loxley, P.~N. (2017).
\newblock The two-dimensional gabor function adapted to natural image statistics: a model of simple-cell receptive fields and sparse structure in images.
\newblock \emph{Neural Comp.}, \emph{29}, 2769--2799.

\bibitem[{MacKay(2003)}]{mackay}
MacKay, D. J. C. (2003).
\newblock \emph{Information theory, inference, and learning algorithms}.
\newblock Cambridge University Press.

\bibitem[{McKenzie et al.(2017)}]{mckenzie}
McKenzie, M., Loxley, P., Billingsley, W., \& Wong, S. (2017).
\newblock Competitive reinforcement learning in Atari games.
\newblock \emph{Peng W., Alahakoon D., Li X. (eds) AI 2017: Advances in Artificial Intelligence. AI 2017. Lecture Notes in Computer Science}, \emph{10400}, Springer.

\bibitem[{Mnih et al.(2015)}]{mnih}
Mnih, V., et al. (2015).
\newblock Human-level control through deep reinforcement learning.
\newblock \emph{Nature}, \emph{518}, 529--533.

\bibitem[{Mumford and Gidas(2001)}]{mumford}
Mumford, D., \& Gidas, B. (2001).
\newblock Stochastic models for generic images.
\newblock \emph{Quarterly Appl. Math.}, \emph{59}, 85--111. 

\bibitem[{Olshausen(2013)}]{olshausen13b}
Olshausen, B.~A. (2013).
\newblock Highly overcomplete sparse coding.
\newblock \emph{Proc.~SPIE 8651, Human Vision and Electronic Imaging XVIII}, \emph{doi:10.1117/12.2013504}.

\bibitem[{Olshausen and Field(1996)}]{olshausen96}
Olshausen, B.~A., \& Field, D.~J. (1996).
\newblock Emergence of simple-cell receptive field properties by learning a sparse code for natural images.
\newblock \emph{Nature}, \emph{381}, 607 -- 609.

\bibitem[{Olshausen and Field(1997)}]{olshausen97}
Olshausen, B.~A., \& Field, D.~J. (1997).
\newblock Sparse coding with an overcomplete basis set: a strategy employed by V1?
\newblock \emph{Vision Res.}, \emph{37}, 3311--3325.

\bibitem[{Papyan et al.(2018)}]{papyan}
Papyan, V., Romano, Y., Sulam, J., \& Elad, M. (2018).
\newblock Theoretical foundations of deep learning via sparse representations: a multilayer sparse model and its connection to convolutional neural networks.
\newblock \emph{IEEE Signal Proc Mag.}, \emph{35}, 72--89.

\bibitem[{Perrinet(2010)}]{perrinet1}
Perrinet, L.~U., (2010).
\newblock Role of homeostasis in learning sparse representations.
\newblock \emph{Neural Comp.}, \emph{22}, 1812--1836.

\bibitem[{Perrinet(2019)}]{perrinet2}
Perrinet, L.~U., (2019).
\newblock An Adaptive Homeostatic Algorithm for the Unsupervised Learning of Visual Features.
\newblock \emph{Vision}, \emph{3}, 47.

\bibitem[{Petrov and Li(2003)}]{petrov}
Petrov, Y., \& Li, Z. (2003).
\newblock Local correlations, information redundancy, and sufficient pixel depth in natural images.
\newblock \emph{J.~Opt.~Soc.~Am.~A}, \emph{20}, 56 -- 66.

\bibitem[{Rehn and Sommer(2007)}]{sommer}
Rehn, M., \& Sommer, T. (2007).
\newblock A network that uses few active neurones to code visual input predicts the diverse shapes of cortical receptive fields.
\newblock \emph{J.~Comput.~Neurosci.}, \emph{22}, 135 -- 146.

\bibitem[{Ruderman and Bialek(1994)}]{ruderman2}
Ruderman, D.~L., \& Bialek, W. (1994).
\newblock Statistics of natural images: scaling in the woods.
\newblock \emph{Phys.~Rev.~Lett.}, \emph{73}, 814--817.

\bibitem[{Ruderman(1997)}]{ruderman}
Ruderman, D.~L. (1997).
\newblock Origins of Scaling in Natural Images.
\newblock \emph{Vision Res.}, \emph{37}, 3385--3398.

\bibitem[{Tibshirani(1996)}]{tib}
Tibshirani, R. (1996). 
\newblock Regression shrinkage and selection via the lasso. 
\newblock \emph{J. R. Statist. Soc. B}, \emph{58}, 267--288.

\bibitem[{Vinje \& Gallant(2000)}]{gallant}
Vinje, W.~E., \& Gallant, J.~L. (2000).
\newblock Sparse coding and decorrelation in primary visual cortex during natural vision.
\newblock \emph{Science}, \emph{287}, 1273--1276.

\end{thebibliography}
\end{document}